# Automated Chest CT Protocol Selection via Large Language Model Derived Text Embeddings from Imaging Request Text

Zahra Hosseini, MSc[1], Mahan Pouromidi, MSc[1], Farzad Khalvati, MSc, PhD[3*], Patrik Rogalla, MD, PhD, MBA[2*]

[1] Joint Department of Medical Imaging (JDMI), University Health Network – Toronto General Hospital, The Hospital for Sick Children, Toronto, Ontario, Canada

[2] Joint Department of Medical Imaging (JDMI), University Health Network – Toronto General Hospital, University of Toronto, Toronto, Ontario, Canada

[3] Department of Medical Imaging, University of Toronto, and The Hospital for Sick Children, Toronto, Ontario, Canada

* Co-senior authors

Corresponding author: Patrik Rogalla, MD, PhD, MBA, Joint Department of Medical Imaging, Toronto General Hospital, 585 University Avenue, Toronto, ON, M5G 2N2, Canada. Email: patrik.rogalla@uhn.ca

## Abstract

**Purpose:** Accurate CT protocol selection is critical for diagnostic quality and patient safety, yet the current process is manual, time-consuming, and prone to inconsistencies. Prior Machine Learning methods using keywords or bag-of-words lack contextual understanding and perform poorly on rare protocols. We propose a decision support system using large language model (LLM) features to recommend protocols from free-text clinical indications, capturing clinical nuance and phrasing variation for more consistent, efficient selection.

**Methods:** In this REB-approved retrospective study, 285,123 chest CT imaging requests from a large academic medical center (2017–2024) were split into training (228,099, 80%) and held-out test (57,024, 20%) sets. Each request included procedure names, clinical indication, HIS comments, and the selected protocol. Clinical text was embedded using a fine-tuned LLM, Meta's LLaMA-3.1-70B; these features input a logistic regression classifier predicting 18 protocol labels (e.g., PE, LDCT).

**Results:** The pipeline achieved a weighted precision of 0.84, weighted F1-score of 0.81, and overall accuracy of 79% across 18 CT protocols. On 300 independent cases with expert consensus, the LLM reached an overall accuracy of 80% versus 83% for radiologists, with no significant difference ($p = 0.263$). Performance was comparable across most classes, with the LLM exceeding radiologists for some challenging categories, and entropy analyses indicated more balanced protocol use, suggesting reduced variability.

**Conclusion:** An LLM-based recommendation system can leverage general knowledge from a large natural-text corpus to accurately assign chest CT protocols from free-text imaging requests, and may serve as a viable foundation for protocol recommendation tools where inputs require language understanding.

## Introduction

Computed tomography (CT) is a cornerstone of diagnostic imaging, with chest CT examinations serving as an essential diagnostic tool for evaluating pulmonary, cardiovascular, and oncologic diseases. Appropriate protocol selection helps provide optimal image quality, most accurate diagnosis, and patient safety by determining scan parameters, radiation dose, and contrast administration.[1,2] Clinically, this selection is performed manually by radiologists (or delegated to technologists) by reviewing the imaging requests, a process that can be time-consuming and variable across radiologists.

Manual protocoling introduces workflow inefficiencies and inconsistency in clinical practice, particularly when free-text requests are ambiguous or incomplete.[3–5] Recent studies have explored natural language processing (NLP) methods to automate protocol selection using the textual content of imaging orders. However, traditional keyword-based and bag-of-words[6,7] approaches capture limited clinical context and often fail to generalize to rare or atypical protocols.[8,9] More recently, large language models have been applied directly to imaging protocol selection, achieving high accuracy for abdominal and pelvic CT using prompt-based context engineering and for CT protocol assignment using a fine-tuned model as a clinical decision-support tool.[10,11]

Large language models (LLMs) such as LLaMA-3 have recently demonstrated strong capabilities in contextual understanding and semantic reasoning across diverse biomedical text tasks.[12] By encoding free-text indications into dense embeddings, these models can capture nuanced relationships between symptoms, risk factors, and imaging requirements. We hypothesized that LLM-derived embeddings could improve the accuracy and consistency of automated chest CT protocol selection compared with traditional NLP baselines. This study evaluates an LLM-based decision-support system for protocol assignment and compares its performance against radiologist-selected protocols across common and rare chest CT protocols.

## Materials and Methods

### *Study Design*

This retrospective study at a large academic medical center was approved by the institutional Research Ethics Board (REB) with a waiver of informed consent. All data handling complied with the jurisdiction's Personal Health Information Protection Act. Chest CT imaging requests received between January 1, 2017, and December 31, 2024, were extracted from the Radiology Information System (RIS) and Hospital Information System (HIS). Each record contained a procedure name, a free-text clinical indication, and optional HIS comments.

The primary objective was to evaluate whether LLM-derived text embeddings improve automated chest CT protocol assignment compared with standard clinical assignments. Secondary objectives were: (1) evaluating performance across common and rare protocol classes, (2) comparing model predictions to expert radiologist consensus in an independent reader study, and (3) demonstrating a PHI-preserving approach for fine-tuning large language models locally within a secure hospital environment rather than relying on cloud-based models.

### *Study Population*

The study population included all patients who underwent chest CT during the study period; inclusion criteria were: (1) at least one free-text clinical indication, (2) a finalized CT protocol assigned at scan time, and (3) availability of both procedure and HIS comment fields. After de-identification and quality control, 285,123 unique imaging requests were retained and divided into 80% training (228,099) and 20% test (57,024) subsets. Patients were 49.9% female and 50.1% male, with a mean age of 63.0 ± 15.1 years (range, 13–106); mean age was 62.6 ± 15.1 years for female and 63.4 ± 15.1 years for male patients. Age and sex were not used as model inputs.

### *Text Preprocessing*

The procedure name, clinical indication, and HIS comments, were concatenated into a single composite string per case. Preprocessing included lower-casing, whitespace removal, and Unicode normalization. Common clinical abbreviations were expanded (e.g., "SOB" to "shortness of breath," "r/o" to "rule out") using the CLEVER terminology resource.[13,14] Dates, identifiers, and unrelated numerical codes were removed to preserve patient privacy. An example of the transformation from raw RIS/HIS fields to the composite text string used for embedding is shown in supplemental Table S1.

Protocol labels were harmonized with board-certified thoracic radiologists into an 18-class ontology covering the full range of chest CT protocols performed at our institution; protocols with fewer than 100 occurrences were not retained (supplemental Table S2). The class distribution exhibited a pronounced long-tail pattern, illustrated in Figure 1, in which a few common protocols dominated while many specialized protocols appeared infrequently.[14]

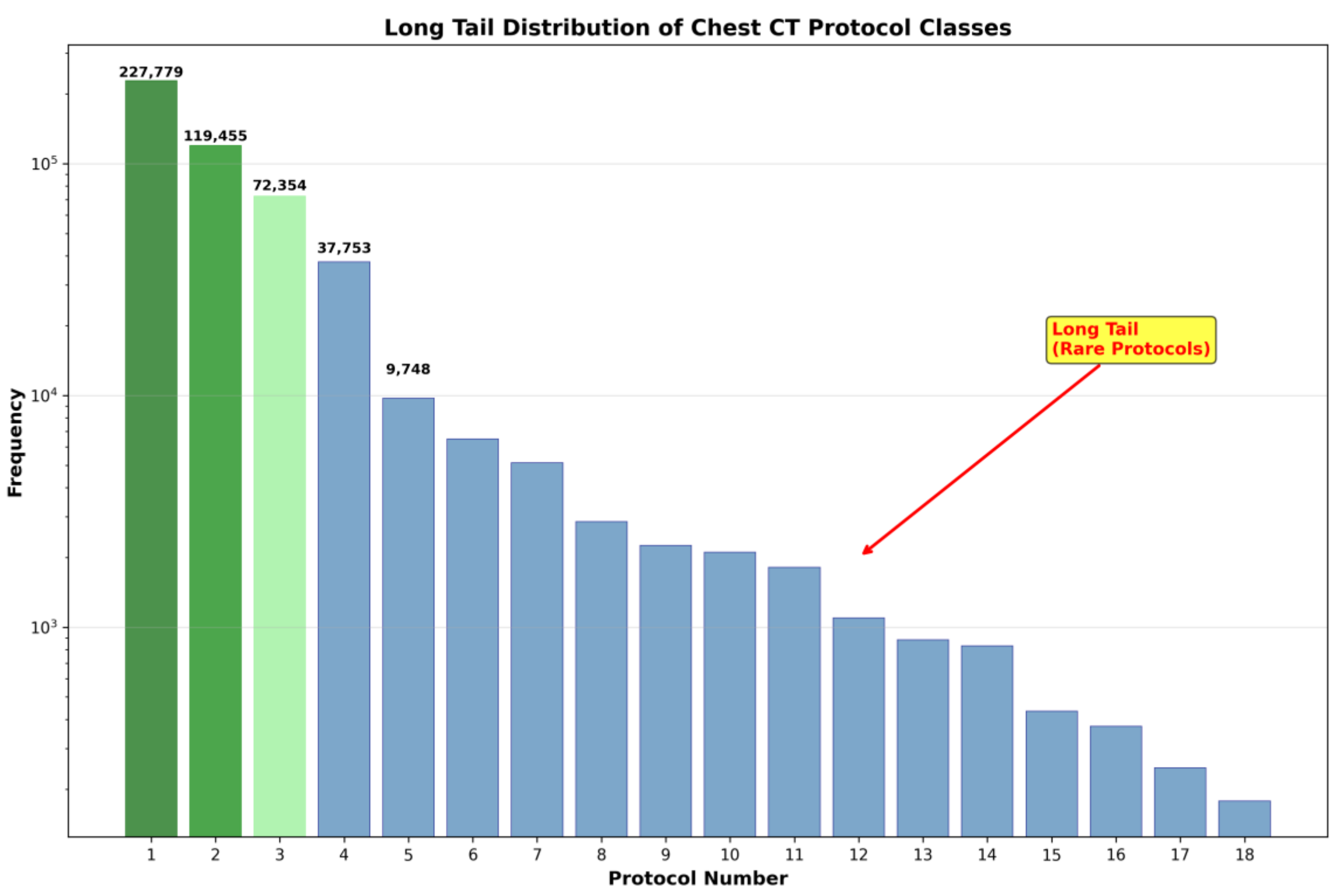

**Figure 1.** Long-tail distribution of chest CT protocols before pre-processing. A few frequent protocols dominate the dataset, while most specialized protocols have limited representation.

### *LLM Fine-Tuning and Embedding Extraction*

We used an open-weight LLaMA-3.1-70B model[15] adapted locally on institution-specific chest CT request text. Fine-tuning used the Unsloth implementation,[16] which supports parameter-efficient adaptation with 4-bit NF4 quantization; the model used `bfloat16` precision where supported, with FlashAttention[17] enabled to reduce memory usage. Low-Rank Adaptation (LoRA)[18] was applied by inserting trainable low-rank matrices into the attention layers while keeping the base model weights frozen (rank $r = 8$, $\alpha = 32$, dropout = 0.05; maximum sequence length 2048 tokens). Training used the AdamW-8bit optimizer with a learning rate of $2\times10^{-5}$, linear scheduling, and weight decay of 0.01, for 1500 steps with evaluation and checkpointing every 250 and 500 steps, respectively.

Embeddings were generated directly from the model's hidden representations. For each input sequence, the final-layer hidden states were extracted and a single fixed-length embedding was computed by mean pooling across the token dimension:

$$\boldsymbol{e} = \text{mean}(\text{HiddenStates}_{\text{layer_final}}) \tag{1}$$

This yields a stable, prompt-independent embedding, avoiding variability from prompt formatting or instruction tuning.

### *Model Training and Class Imbalance Handling*

The embedding vectors were used to train a multinomial logistic regression classifier with L2 regularization and class-weighted cross-entropy loss. Given the extreme class imbalance, all protocols with fewer than 2,000 examples were upsampled by repetition to reach the target count while preserving the original label distribution. Of the training set, 10% was held out for validation; convergence was monitored using macro-F1, and the regularization coefficient was selected via grid search over $C \in \{0.01, 0.1, 1, 10\}$.

### *Implementation Environment*

All training and inference ran on institutional high-performance computing infrastructure (16 CPU cores, 256 GB RAM, one NVIDIA H100 80 GB GPU) under Slurm, using Python 3.10, PyTorch 2.2, Hugging Face Transformers v4.40, `bitsandbytes`, and Unsloth.

### *Evaluation Metrics*

Model predictions on the independent test set were compared against ground-truth protocol labels using overall accuracy, weighted precision, weighted recall, weighted F1-score, and macro-averaged F1-score. Performance on the lowest-frequency quartile of protocols (rare classes) was analyzed separately to assess robustness to long-tail class imbalance.

In addition, an expert reader study was conducted on 300 cases. Four experienced radiologists independently assigned a protocol for each case using the available clinical information, patient history, and imaging requirements.

Statistical comparisons between the model-generated recommendations and the protocols assigned during routine clinical workflow, each evaluated against the radiologist consensus reference standard, were performed using McNemar's test for paired accuracy[19] and paired *t*-tests on per-class F1-scores,[19,20] with significance assessed at $\alpha = 0.05$.

## Results

A total of 285,123 chest CT imaging requests met the inclusion criteria during the study period, comprising a training set of 228,099 (80.0%) and an independent test set of 57,024 (20.0%).

Overall test set performance (Table 1) shows that the model achieves high precision and recall across 18 protocol classes. Unlike the training set, which underwent upsampling for class balance, the test set consisted entirely of real, non-augmented data (57,024 cases), providing an unbiased evaluation of model performance.

**Table 1.** Overall Performance on the Held-Out Test Set (20%)

| Metric | Precision | Recall | F1 Score |
|---|---|---|---|
| Overall | 0.847 | 0.79 | 0.81 |
| **Support** | | 57,024 | |

**Table 2.** Per-Class Accuracy for Clinical Protocols vs. Radiologist Consensus

| # | Protocol Class | Correct | Total | Accuracy |
|---|---|---|---|---|
| 1 | Aorta – dissection, chest+abdomen | 10 | 11 | 0.90 (90%) |
| 2 | Aorta – dissection, chest only | 8 | 10 | 0.80 (80%) |
| 3 | Esophageal Cancer (R + C + oral) | 14 | 14 | 1.00 (100%) |
| 4 | ILD baseline (R + HighRes + exp + prone) | 8 | 8 | 1.00 (100%) |
| 5 | Low Dose CT (LDCT) | 20 | 21 | 0.95 (95%) |
| 6 | Lung Transplant (LDCT + exp) | 11 | 14 | 0.78 (78%) |
| 7 | PE (CTPA) | 14 | 18 | 0.77 (77%) |
| 8 | PE + R + C | 9 | 12 | 0.75 (75%) |
| 9 | PE + dissection, chest only | 10 | 10 | 1.00 (100%) |
| 10 | PE – ILD (LDCT + PE) | 8 | 8 | 1.00 (100%) |
| 11 | PE – iodine subtraction | 10 | 11 | 0.90 (90%) |
| 12 | R | 59 | 69 | 0.85 (85%) |
| 13 | R + C | 41 | 65 | 0.63 (63%) |
| 14 | SVCO (bilateral contrast) | 9 | 9 | 1.00 (100%) |
| 15 | Trachea (R + thin section for 3D) | 9 | 9 | 1.00 (100%) |
| 16 | Trauma – Chest | 10 | 11 | 0.90 (90%) |
| | **Overall accuracy** | 250 | 300 | **0.83 (83%)** |

The 300-case evaluation set initially contained at least 10 cases for each of the 18 protocols. After the four-reader consensus reference standard was established, all cases historically assigned to two protocols, Lung Cancer (R + C, plus abdomen) and Mediastinal Mass (iodine subtraction), were assigned an alternative protocol by every reader, leaving those two protocols with no cases under the reference standard; four further protocols (Trachea, PE–ILD, ILD baseline, and SVCO) lost 1, 2, 2, and 1 cases, respectively, because the consensus label differed from the historically assigned protocol. The reader-study analysis therefore spans 16 protocol classes across all 300 cases, consistent with the prior report on this evaluation set.[14] Table 2 presents the per-class accuracy for radiologist-assigned protocols when evaluated against the established consensus. The overall accuracy was 83% (250/300).

Against this consensus, the LLM-based classifier achieved an overall accuracy of 80% (241/300), compared with 83% (250/300) for human.

**Table 3.** Per-Class Accuracy for LLM Predictions Compared to Expert Consensus (4 Radiologists)

| # | Protocol Class | Correct | Total | Accuracy |
|---|---|---|---|---|
| 1 | Aorta – dissection, chest+abdomen | 9 | 11 | 0.82 (82%) |
| 2 | Aorta – dissection, chest only | 8 | 10 | 0.80 (80%) |
| 3 | Esophageal Cancer (R + C + oral) | 14 | 14 | 1.00 (100%) |
| 4 | ILD baseline (R + HighRes + exp + prone) | 8 | 8 | 1.00 (100%) |
| 5 | Low Dose CT (LDCT) | 19 | 21 | 0.90 (90%) |
| 6 | Lung Transplant (LDCT + exp) | 13 | 14 | 0.92 (92%) |
| 7 | PE (CTPA) | 15 | 18 | 0.83 (83%) |
| 8 | PE + R + C | 4 | 12 | 0.33 (33%) |
| 9 | PE + dissection, chest only | 7 | 10 | 0.70 (70%) |
| 10 | PE – ILD (LDCT + PE) | 7 | 8 | 0.87 (87%) |
| 11 | PE – iodine subtraction | 9 | 11 | 0.82 (82%) |
| 12 | R | 62 | 69 | 0.89 (89%) |
| 13 | R + C | 43 | 65 | 0.66 (66%) |
| 14 | SVCO (bilateral contrast) | 7 | 9 | 0.78 (78%) |
| 15 | Trachea (R + thin section for 3D) | 9 | 9 | 1.00 (100%) |
| 16 | Trauma – Chest | 7 | 11 | 0.63 (63%) |
| | **Overall accuracy** | 241 | 300 | **0.80 (80%)** |

***Statistical Significance***

Against the consensus reference standard, the protocol assigned in routine clinical workflow achieved an accuracy of 83.33% (250/300), whereas the LLM-based classifier achieved an accuracy of 80.33% (241/300). McNemar's test showed no statistically significant difference between the two approaches ($\chi^2 = 1.25$, $p = 0.263$). There were 21 cases in which the LLM matched the consensus while the routine clinical assignment did not, and 30 cases in which the routine clinical assignment matched the consensus while the LLM did not.

**Table 4.** McNemar's Test Results: Human vs. LLM Protocol Assignment Accuracy

| Performance Category | Count | Percentage |
|---|---|---|
| Both correct | 220 | 73.3% |
| Both incorrect | 29 | 9.7% |
| Human correct, LLM incorrect | 30 | 10.0% |
| Human incorrect, LLM correct | 21 | 7.0% |
| **Total discordant pairs** | **51** | **17.0%** |
| **Chi-squared statistic ($\chi^2$)** | **1.25** | |
| **P-value** | **0.263** | |

Figure 2 compares the Shannon entropy of protocol distributions between the LLM and individual radiologists.[21] Shannon entropy quantifies the distributional diversity of protocol assignments, where lower values indicate that predictions are concentrated on a limited set of common protocols, and higher values reflect a more balanced use of the full range of available options. The LLM exhibited higher entropy ($H = 3.747$) compared to all radiologists ($H = 3.449$–3.555), indicating that its predictions were more evenly distributed across protocol classes.

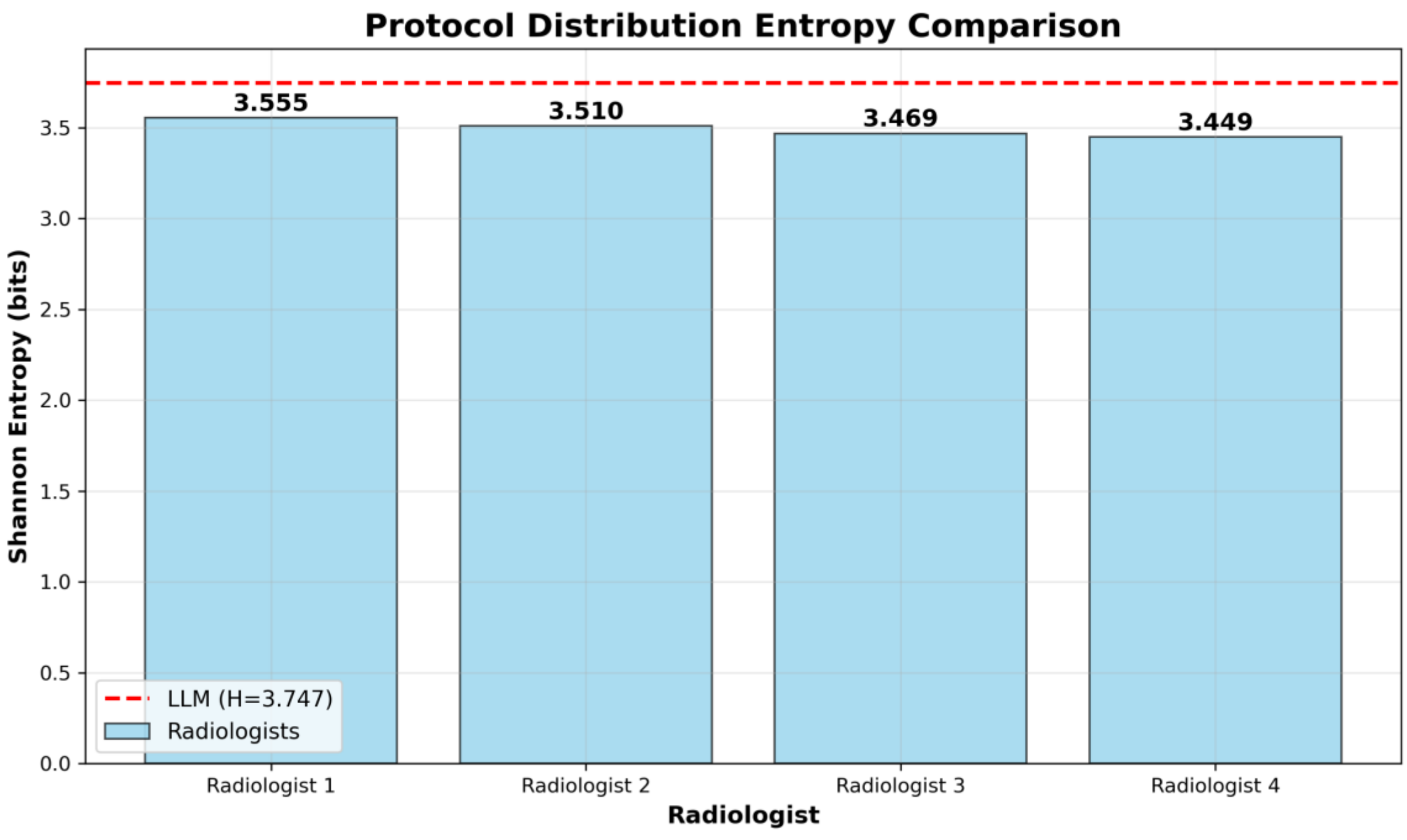

**Figure 2.** Comparison of Shannon entropy (H, bits) for protocol assignment distributions between the LLM-based classifier and four individual board-certified radiologists. The y-axis represents Shannon

entropy, which quantifies distributional diversity of protocol assignments; lower values indicate predictions concentrated on a limited subset of common protocols, while higher values reflect more balanced use of the full range of 18 protocol classes. The LLM (dashed red reference line, H = 3.747) exhibited higher entropy than all four radiologists (H = 3.449–3.555), indicating more even distribution of protocol assignments across classes and a reduced tendency toward over-selection of common protocols.

## Discussion

Accurate CT protocol selection is a critical step in radiology workflow, directly influencing diagnostic quality, radiation exposure, contrast timing, and downstream management.[22] In routine practice, protocol assignment relies on the radiologist's ability to interpret often brief or ambiguous free-text requests and map them to an increasingly complex set of tailored protocols. Even among experienced readers, variability in protocol selection is well documented, reflecting the subjective, context-dependent nature of interpreting clinical indications.[23] We evaluated whether a large language model (LLM)-based system could support this process with consistent, clinically aligned recommendations derived from free-text order information.

Our results show the proposed system performs comparably to expert thoracic radiologists. Against a four-reader consensus reference standard, the model achieved an overall accuracy of 80.3%, closely matching radiologist performance at 83.3%; the difference was not statistically significant, indicating that its decisions were largely indistinguishable from those of human experts. This suggests LLM-derived embeddings capture clinically meaningful nuances in request text, letting the model approximate the reasoning radiologists apply when selecting protocols.

Class-level analysis highlights the complementary strengths of experts and the LLM. Both performed uniformly well on protocols with clear, specific indications, such as baseline ILD assessment, esophageal cancer staging, and tracheal imaging, where clinical intent is typically unambiguous. The LLM showed comparatively stronger performance on classes with broader or more heterogeneous presentations; for example, it exceeded radiologists on the standard noncontrast chest protocol (R) and lung transplant follow-up, both of which span diverse clinical contexts and benefit from stable interpretation of textual cues.

Conversely, the model underperformed on a subset of rare, high-complexity protocols, including Trauma–Chest and combined PE + R + C (early perfusion phase for PE detection and parenchymal phase for tissue contrast). These represented fewer than 1% of cases, and their scarcity likely limited fine-grained learning. This long-tail challenge is well recognized in clinical practice and machine learning: radiologists manage such cases through years of exposure to unusual presentations, whereas data-driven models depend on sufficient examples to generalize. Improving this regime may require targeted data augmentation, rule-based safeguards, or protocol-specific fine-tuning to reinforce decision boundaries that human readers internalize through experience.

Beyond accuracy, the entropy analysis showed the LLM distributed assignments more evenly than individual radiologists, suggesting a reduced tendency to over-select common protocols. This is valuable in high-volume settings where subtle differences may be missed under time pressure or alert fatigue, and may help standardize practice and reduce inter-reader variability.[24]

The system is intended as a decision-support tool that streamlines routine steps and reduces cognitive load. Because most worklist cases follow familiar patterns, accurate automated triage of standard cases can help technologists prepare studies efficiently and let radiologists focus on complex or ambiguous requests.[25] Such support may be especially valuable in resource-constrained settings, after-hours coverage, and training environments where experience varies.

Fine-tuning a large open-weight model entirely within the institution's secure computing environment further strengthens this approach. Because no patient information leaves hospital infrastructure, the workflow respects privacy while adapting LLMs to local clinical language and institutional protocol structure. This is particularly relevant in jurisdictions where imaging request text cannot be transmitted to external services because of privacy and data-governance requirements. As departments increasingly adopt AI tools, this study may provide a blueprint for integrating LLMs safely and effectively into existing systems.

This study has several limitations. It was a single-center, retrospective study, so external multi-institutional validation is needed before generalization. Performance remained limited on the rarest protocols (e.g., PE + R + C, 33%; Trauma–Chest, 63%), reflecting the few training examples available for these classes. Finally, the system was not tested prospectively, where its effect on workflow and patient outcomes remains to be established.

In summary, an LLM-based protocol recommendation system can achieve radiologist-level performance while improving consistency across a broad range of chest CT protocols. Although certain rare protocols remain challenging and warrant refinement, the results demonstrate the potential of LLM-driven decision support to reduce variability, improve efficiency, and improve the reliability of protocol selection in practice. Protocol recommendation may be a natural extension of the foundation-model paradigm in radiology, a shift supported by a recent survey[26] that emphasizes developing such task-specific applications while addressing deployment risks, interpretability, and performance in diverse settings.

**Acknowledgments**

**Conflict of Interest:** The authors declare no conflicts of interest related to the methodology, data, or tools used in this study. No commercial entity provided funding or had any role in the study design, data collection, analysis, interpretation, or manuscript preparation.

**REB/PHIPA Approval:** This study was approved by the institutional Research Ethics Board (REB) with a waiver of individual informed consent due to the retrospective nature of the work and the use of de-identified data. All data handling complied with the Personal Health Information Protection Act (PHIPA) of Ontario and equivalent provincial and federal privacy legislation in Canada. All analyses were performed on secure, access-controlled servers within the institution's research computing environment.

**Consent:** Written informed consent was waived by the REB due to the use of retrospective, de-identified data and the absence of direct patient contact.

## References

1. Little BP. Approach to Chest Computed Tomography. Clinics in Chest Medicine. 2015;36(2):127–145.
2. Hsieh MS, Chiu CS, How CK, et al. Contrast Medium Exposure During Computed Tomography and Risk of Development of End-Stage Renal Disease in Patients With Chronic Kidney Disease: A Nationwide Population-Based, Propensity Score-Matched, Longitudinal Follow-Up Study. Medicine. 2016;95(16):e3388.

3. Kubo T, Lin PJP, Stiller W, et al. Radiation Dose Reduction in Chest CT: A Review. American Journal of Roentgenology. 2008;190(2):335–343.
4. Mahesh M. Variability in CT Protocols. Journal of the American College of Radiology. 2013;10(10):805–806.
5. Nencka AS, Sherafati M, Goebel T, Tolat P, Koch KM. Deep-Learning Based Tools for Automated Protocol Definition of Advanced Diagnostic Imaging Exams. ArXiv. 2021;2106.08963 [preprint]. https://arxiv.org/abs/2106.08963. Posted May 28, 2021. Accessed September 4, 2026.
6. Sparck Jones K. A Statistical Interpretation of Term Specificity and Its Application in Retrieval. Journal of Documentation. 1972;28(1):11–21.
7. Salton G, Wong A, Yang CS. A Vector Space Model for Automatic Indexing. Communications of the ACM. 1975;18(11):613–620.
8. Xavier BA, Chen PH. Natural Language Processing for Imaging Protocol Assignment: Machine Learning for Multiclass Classification of Abdominal CT Protocols Using Indication Text Data. Journal of Digital Imaging. 2022;35(5):1120–1130.
9. Kalra A, Chakraborty A, Fine B, Reicher J. Machine Learning for Automation of Radiology Protocols for Quality and Efficiency Improvement. Journal of the American College of Radiology. 2020;17(9):1149–1158.
10. Buckley BW, Dias AB, Deng Y, et al. Optimizing Large Language Models for Automated Protocoling of Abdominal and Pelvic CT Scans: The Power of Context. Radiology. 2026;318(1):e252105.
11. Kanemaru N, Yasaka K, Okimoto N, et al. Efficacy of Fine-Tuned Large Language Model in CT Protocol Assignment as Clinical Decision-Supporting System. Journal of Imaging Informatics in Medicine. 2025;38(6):4336–4348.
12. Singhal K, Azizi S, Tu T, et al. Large Language Models Encode Clinical Knowledge. Nature. 2023;620(7972):172–180.
13. Tamang S. CLEVER (CLinical EVEnt Recognizer). GitHub website. https://github.com/stamang/CLEVER. Accessed August 11, 2026.
14. Rogalla P, Fratesi J, Kandel S, Patsios D, Khalvati F, Carey S. Development and Evaluation of an Automated Protocol Recommendation System for Chest CT Using Natural

Language Processing With CLEVER Terminology Word Replacement. Canadian Association of Radiologists Journal. 2025;76(2):257–264.

15. Grattafiori A, Dubey A, Jauhri A, et al. The Llama 3 Herd of Models. ArXiv. 2024;2407.21783 [preprint]. https://arxiv.org/abs/2407.21783. Posted July 31, 2024. Updated November 23, 2024. Accessed September 4, 2026.
16. Han D, Han M, Unsloth team. Unsloth. GitHub website. https://github.com/unslothai/unsloth. Accessed August 11, 2026.
17. Dao T, Fu D, Ermon S, Rudra A, Ré C. FlashAttention: Fast and Memory-Efficient Exact Attention with IO-Awareness. Adv Neural Inf Process Syst. 2022;35:16344–16359.
18. Hu EJ, Shen Y, Wallis P, et al. LoRA: Low-Rank Adaptation of Large Language Models. ArXiv. 2021;2106.09685 [preprint]. https://arxiv.org/abs/2106.09685. Posted June 17, 2021. Updated October 16, 2021. Accessed September 4, 2026.
19. McNemar Q. Note on the Sampling Error of the Difference Between Correlated Proportions or Percentages. Psychometrika. 1947;12(2):153–157.
20. Meystre SM, Savova GK, Kipper-Schuler KC, Hurdle JF. Extracting Information from Textual Documents in the Electronic Health Record: A Review of Recent Research. Yearbook of Medical Informatics. 2008;17(1):128–144.
21. Shannon CE. A Mathematical Theory of Communication. Bell System Technical Journal. 1948;27(3):379–423.
22. Denck J, Haas O, Guehring J, Maier A, Rothgang E. Automated Protocoling for MRI Exams—Challenges and Solutions. Journal of Digital Imaging. 2022;35(5):1293–1302.
23. Schemmel A, Lee M, Hanley T, et al. Radiology Workflow Disruptors: A Detailed Analysis. Journal of the American College of Radiology. 2016;13(10):1210–1214.
24. Wong KA, Hatef A, Ryu JL, Nguyen XV, Makary MS, Prevedello LM. An Artificial Intelligence Tool for Clinical Decision Support and Protocol Selection for Brain MRI. American Journal of Neuroradiology. 2023;44(1):11–16.
25. Ranschaert E, Topff L, Pianykh O. Optimization of Radiology Workflow with Artificial Intelligence. Radiologic Clinics of North America. 2021;59(6):955–966.
26. Khan W, Leem S, See KB, Wong JK, Zhang S, Fang R. A Comprehensive Survey of Foundation Models in Medicine. IEEE Reviews in Biomedical Engineering. 2026;19:283–304.

## Supplemental Material

**Table S1.** Example of raw vs. preprocessed imaging request text after normalization and field harmonization.

| Original order text (HIS/RIS fields) | Preprocessed text used for LLM |
|---|---|
| Indication: bronchiectasis<br>HIS: Area: Lung // Comment: inspiratory and expiratory phases to look for bronchiolitis obliterans post lung transplant // Hx of renal disease?: yes // Last creatinine result: 115 (umol/L) // Ordering physician: – // Patient diabetic?: yes | bronchiectasis, inspiratory and expiratory phases to evaluate bronchiolitis obliterans after lung transplant, diabetic, history of renal disease, last creatinine 115 µmol/L |

HIS = Hospital Information System; RIS = Radiology Information System; LLM = large language model.

**Table S2.** Frequency and relative frequency of chest CT protocol classes before preprocessing. The protocols are sorted by frequency.

| # | Protocol | Frequency | Percent (%) |
|---|---|---|---|
| 1 | R | 227,779 | 46.3 |
| 2 | R + C | 119,455 | 24.3 |
| 3 | Low Dose CT (LDCT) | 72,354 | 14.7 |
| 4 | PE (CTPA) | 37,753 | 7.7 |
| 5 | Lung Transplant (LDCT + exp) | 9,748 | 2.0 |
| 6 | ILD baseline (R + HighRes + exp + prone) | 6,481 | 1.3 |
| 7 | Esophageal Cancer (R + C + oral) | 5,129 | 1.0 |
| 8 | PE + R + C | 2,850 | 0.58 |
| 9 | Lung Cancer (R + C, plus abdomen) | 2,257 | 0.46 |
| 10 | Aorta – dissection (chest only) | 2,103 | 0.43 |
| 11 | Aorta – dissection (chest + abdomen) | 1,814 | 0.37 |
| 12 | PE – ILD (LDCT + PE) | 1,098 | 0.22 |
| 13 | PE – iodine subtraction | 881 | 0.18 |
| 14 | PE + dissection (chest only) | 830 | 0.17 |
| 15 | Trachea (R + thin section for 3D) | 434 | 0.089 |
| 16 | Trauma – Chest | 374 | 0.076 |
| 17 | SVCO (bilateral contrast) | 247 | 0.050 |
| 18 | Mediastinal Mass (iodine subtraction) | 178 | 0.036 |